\documentclass{article}
\usepackage[T1]{fontenc}
\usepackage[utf8]{inputenc}
\usepackage[margin=1in]{geometry}
\usepackage{authblk}
\usepackage{hyperref}
\usepackage{tikz}
\usepackage{pgfplots}
\usepackage{tikz-dependency}
\usepackage{comment}
\usepackage{lmodern}
\usepackage{array}
\usepackage{color}    
\usepackage{graphicx}
\usepackage{multirow}
\usepackage{tabularx}
\usepackage{upquote}  
\usepackage{beramono} 
\usepackage{listings} 
\title{Yara Parser: A Fast and Accurate Dependency Parser}
\author[1]{Mohammad Sadegh Rasooli\thanks{Implementation of the core structure of the parser was done during a summer internship at Yahoo Labs NYC.}}
\author[2]{Joel Tetreault}
\affil[1]{~Department of Computer Science, Columbia University, New York, NY, \textit{rasooli@cs.columbia.edu}}
\affil[2]{~Yahoo Labs, New York, NY, \textit{tetreaul@yahoo-inc.com}}
\date{March 2015}

\begin{document}
  \maketitle

  \abstract{
  Dependency parsers are among the most crucial tools in natural language processing as they have many important applications in downstream tasks such as information retrieval, machine translation and knowledge acquisition. We introduce the Yara Parser, a fast and accurate open-source dependency parser based on the arc-eager algorithm and beam search.  It achieves an unlabeled accuracy of 93.32 on the standard WSJ test set which ranks it among the top dependency parsers.  At its fastest, Yara can parse about 4000 sentences per second when in greedy mode (1 beam).  When optimizing for accuracy (using 64 beams and Brown cluster features), Yara can parse 45 sentences per second.
The parser can be trained on any syntactic dependency treebank and different options are provided in order to make it more flexible and tunable for specific tasks. It is released with the Apache version 2.0 license and can be used for both commercial and academic purposes. The parser can be found at {\footnotesize \url{https://github.com/yahoo/YaraParser}}. 
  }
  
\tableofcontents

 \section{Introduction}
Dependency trees are one of the main representations used in the syntactic analysis of sentences. They show explicit syntactic dependencies among words in the sentence \cite{kubler2009dependency}. Many dependency parsers have been released in the past decade. Among them, graph-based and transition-based parsing are two main approaches towards dependency parsing. In graph-based models, the parser aims to find the most likely tree from all possible trees by using maximum spanning tree algorithms often in conjunction with dynamic programming. 
 On the other hand, in transition-based models, a tree is converted to a set of incremental actions and the parser decides to commit an action depending on the current configuration of the partial tree. 
 Graph-based parsers can achieve state-of-the-art performance with the guarantee of recovering the best possible parse, but usually at the expense of speed. On the other hand, transition-based parsers are fast because the parser can greedily choose an action in each configuration and thus it can use arbitrary non-local features to compensate the lack of optimality. 
Also, it is easy to augment the set of actions to extend the functionality of the parser on such tasks as disfluency detection \cite{rasooli-tetreault:2013:EMNLP,rasooli-tetreault:2014:EACL2014-SP,honnibalnon} and punctuation prediction \cite{zhang-EtAl:2013:ACL20132}. They are mostly used in supervised tasks but in rare cases they are also used in unsupervised tasks either with little manual linguistic knowledge or with no prior knowledge \cite{daume2009unsupervised,rasooli-faili:2012:ROBUS-UNSUP2012}.
 
 In this report, we provide a brief introduction to our newly released dependency parser. We show that it can achieve a very high accuracy on the standard English WSJ test set and show that it is very fast even in its slowest mode while getting results very close to state-of-the-art. The structure of this report is as follows: in \S\ref{sec:usage} we provide some details about using Yara both in command line and as an API. We provide technical details about it in \S\ref{sec:yara} and experiments are conducted in \S\ref{sec:experiments}. Finally we conclude in \S\ref{sec:conclusion}.

\section{Using Yara in Practice\label{sec:usage}}
\lstset{language=Java,captionpos=b,tabsize=3,frame=lines,keywordstyle=\color{blue},commentstyle=\color{blue},stringstyle=\color{magenta},numbers=left,numberstyle=\tiny,numbersep=5pt,breaklines=true,showstringspaces=false,basicstyle=\footnotesize,emph={label}}

In this section, we give a brief overview of training the parser, using it from the command-line and also as an API. Finally we introduce a simple NLP pipeline that can parse text files. All technical details for the parser are provided in \S \ref{sec:yara}. The default settings for Yara are expected to be the best in practice for accuracy (except the number of training iterations which is dependent on the data and feature settings).

\subsection{Data format} 
Yara uses the CoNLL 2006 dependency format\footnote{\url{http://ilk.uvt.nl/conll/\#dataformat}}
for training as well as testing.  The CoNLL format is a tabular one in which each word (and its
information) in a sentence occupies one line and sentences are separated by a blank line.
Each line is organized into the following tab-delimited columns: 1) word number (starting at one), 2) word form, 3) word lemma, 4) coarse-grained POS tag, 5) fine-grained POS (part-of-speech) tag, 6) unordered set of syntactic and/or morphological features, separated by a vertical bar (|), or an underscore if not available,  7) head of current token (an integer showing the head number where 0 indicates root token), 8) dependency label, 9) projective head (underscore if not available) and 10) projective dependency labels (underscore if not available).  Blank fields are
represented by an underscore.  Yara only uses the first, second, fourth, seventh and eights columns. 

\subsection{Training and Model Selection}
The jar file in the package can be directly used to train a model with the following command line (run from the root directory of the project):

\texttt{\footnotesize \\>> java -jar jar/YaraParser.jar train -train-file [train-file] -dev [dev-file] -model [model-file] -punc [punc-file]\\}

where \texttt{\footnotesize [train-file]} and \texttt{\footnotesize [dev-file]} are CoNLL files for training and development data and \texttt{\footnotesize [model-file]} is the output path for the trained model file. \texttt{\footnotesize [punc-file]} contains a list of POS tags for punctuations in the treebank (see \S \ref{sec:punc}). The model for each iteration will be saved with the pattern \texttt{\footnotesize [model-file]\_iter[iter\#]}; e.g. \texttt{\footnotesize model\_iter2}. In this way, the user can track the best performing model and delete all others. For cases where there is no development data, the user can remove the \texttt{\footnotesize -dev} option from the command line and use any of the saved model files as the final model based on his/her prior knowledge (15 is a reasonable number).

The other options are as follows:
\begin{itemize}
\item  {\bf -cluster [cluster-file]} Brown cluster file: at most 4096 clusters are supported by Yara (default: empty). The format should be the same as {\footnotesize \url{https://github.com/percyliang/brown-cluster/blob/master/output.txt}}

\item {\bf beam:[beam-width]}; e.g. beam:16 (default is 64).

\item {\bf iter:[training-iterations]}; e.g. iter:10 (default is 20).

\item {\bf unlabeled} (default: labeled parsing, unless explicitly put `unlabeled')

\item {\bf lowercase} (default: case-sensitive words, unless explicitly put `lowercase')

\item {\bf basic} (default: use extended feature set, unless explicitly put `basic')

\item {\bf static} (default: use dynamic oracles, unless explicitly put `static' for static oracles)

\item {\bf early} (default: use max violation update, unless explicitly put `early' for early update)

\item {\bf random} (default: choose maximum scoring oracle, unless explicitly put `random' for randomly choosing an oracle)

\item {\bf nt:\#threads}; e.g. nt:4 (default is 8).

\item {\bf root\_first} (default: put ROOT in the last position, unless explicitly put `root\_first')

\end{itemize}

\subsubsection{Punctuation Files}
\label{sec:punc}
In most dependency evaluations, punctuation symbols and their incoming arcs are ignored. Most parser do this by using hard-coded rules for punctuation attachment. Yara instead allows the user to specify which punctuation POS tags are important to their task by providing a path for a punctuation file (\texttt{\footnotesize [punc-file]}) with the \texttt{\footnotesize -punc} option (e.g. \texttt{\footnotesize -punc punc\_files/wsj.puncs}). If no file is provided, Yara uses WSJ punctuations. The punctuation file contains a list of punctuation POS tags, one per line. The Yara git repository provides punctuation files for WSJ data and Google universal POS tags \cite{petrov2011universal}.

\subsubsection{Some Examples}
Here we provide examples for training Yara with different settings. Essentially we pick those examples where, we think, would be useful in practice.

\paragraph{Training with Brown clusters} This can be done via the \texttt{\footnotesize -cluster} option.

\texttt{\footnotesize \\>> java -jar jar/YaraParser.jar train -train-file [train-file] -dev [dev-file] -model [model-file] -punc [punc-file] -cluster [cluster-file]\\}

\paragraph{Training with the fastest mode} This can be done via the \texttt{\footnotesize basic} and \texttt{\footnotesize beam:1} options.

\texttt{\footnotesize \\>> java -jar jar/YaraParser.jar train -train-file [train-file] -dev [dev-file] -model [model-file] -punc [punc-file] beam:1 basic\\}

\paragraph{Changing the number of iterations} This can be done via the \texttt{\footnotesize iter} option. In the following example, we selected 10 iterations.

\texttt{\footnotesize \\>> java -jar jar/YaraParser.jar train -train-file [train-file] -dev [dev-file] -model [model-file] -punc [punc-file] iter:10\\}

\paragraph{Extending memory consumption} It is possible that Java default setting for memory is less than what is really needed in some particular data sets. In those cases, we can extend the memory size by the JVM \texttt{\footnotesize -Xmx} option. In the following example, memory is extended to ten gigabytes.

\texttt{\footnotesize \\>> java -Xmx10g -jar jar/YaraParser.jar train -train-file [train-file] -dev [dev-file] -model [model-file] -punc [punc-file]\\}

\paragraph{Using very specific options} The following example shows a specific case where Yara trains a model on the training data (\texttt{\footnotesize data/train.conll}), develops it on the development data (\texttt{\footnotesize data/dev.conll}), saves each model in the model file  (\texttt{\footnotesize model/train.model}) for each iteration (\texttt{\footnotesize model/train.model\_iter1}, \texttt{\footnotesize model/train.model\_iter2}, \texttt{\footnotesize model/train.model\_iter3}, etc), uses its specific punctuation list (\texttt{\footnotesize punc\_files/my\_lang.puncs}), uses its specific Brown cluster data (\texttt{\footnotesize data/cluster.path}), trains the model in 10 iterations, with 16 beams and 4 threads and uses \texttt{\footnotesize static} oracle and early update. This is all  done after all words are lowercased (with the \texttt{\footnotesize lowercase} option).

\texttt{\footnotesize \\>> java -Xmx10g -jar jar/YaraParser.jar train -train-file data/train.conll -dev data/dev.conll -model model/train.model -punc punc\_files/my\_lang.puncs -cluster data/cluster.path beam:16 iter:10 unlabeled lowercase static early nt:4 root\_first\\}

\subsection{Test and Evaluation}
The test file can be either a CoNLL file or a POS tagged file.
The output will be a file in CoNLL format.

\paragraph{Parsing a CoNLL file}

\texttt{\footnotesize \\>> java -jar jar/YaraParser.jar parse\_conll -input [test-file] -out [output-file] -model [model-file]\\}

\paragraph{Parsing a tagged file}
The tagged file is a simple file where words and tags are separated by a delimiter (default is underscore). The user can use the option \texttt{\footnotesize -delim [delimiter]} (e.g. \texttt{\footnotesize -delim /}) to change the delimiter. The output will be in CoNLL format.

\texttt{\footnotesize \\>> java -jar jar/YaraParser.jar parse\_tagged -input [test-file] -out [output-file] -model [model-file]\\}

\paragraph{Evaluation} 
Both \texttt{\footnotesize [gold-file]} and \texttt{\footnotesize [parsed-file]} should be in CoNLL format. 

\texttt{\footnotesize \\>> java -jar YaraParser.jar eval -gold [gold-file] -parse [parsed-file] -punc [punc-file]\\}

A more descriptive end-to-end example by using a small amount of German training data\footnote{\url{https://github.com/yahoo/YaraParser/tree/master/sample_data}} is shown in Yara's Github repository. This example is shown at {\footnotesize \url{https://github.com/yahoo/YaraParser\#example-usage}}.

\subsection{Parsing a Partial Tree}
\begin{figure*}[!t]
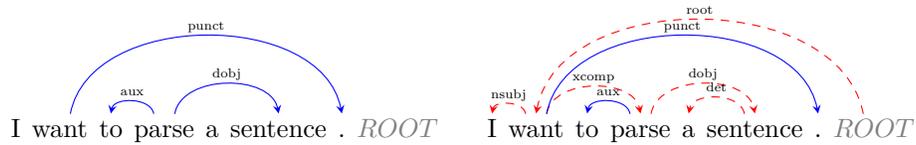

\centering
\begin{dependency}[theme = simple]
\begin{deptext}
\& I \& want \& to \& parse \& a \& sentence \& . \& $\textcolor{gray}{ROOT}$  \\
\end{deptext}
\depedge[color=blue]{5}{4}{aux}
\depedge[color=blue]{5}{7}{dobj}
\depedge[color=blue]{3}{8}{punct}
\end{dependency}
\begin{dependency}[theme = simple]
\begin{deptext}
\& I \& want \& to \& parse \& a \& sentence \& . \& $\textcolor{gray}{ROOT}$  \\
\end{deptext}
\depedge[color=red,dashed]{9}{3}{root}
\depedge[color=red,dashed]{3}{2}{nsubj}
\depedge[color=blue]{5}{4}{aux}
\depedge[color=red,dashed]{3}{5}{xcomp}
\depedge[color=red,dashed]{5}{7}{dobj}
\depedge[color=red,dashed]{7}{6}{det}
\depedge[color=blue]{3}{8}{punct}
\end{dependency}

\caption{\label{fig_partial_tree}A sample partial dependency tree on the left side and its filled tree on the right. As shown in this figure, the added arcs are completely consistent with the partial tree arcs.}
\end{figure*}

Yara can parse partial trees where some gold dependencies are provided and it is expected to return a dependency tree consistent with the partial dependencies. Unknown dependencies are represented with ``-1'' as the head in the CoNLL format. Figure \ref{fig_partial_tree} shows an example of partial parse tree before and after doing constrained parsing.

\texttt{\footnotesize \\>> java -jar YaraParser.jar parse\_partial -input [test-file] -out [output-file] -model [model-file]\\}

\subsection{Yara Pipeline}
We also provide an easy pipeline to use Yara in real applications. The pipeline benefits from the OpenNLP\footnote{\footnotesize \url{http://opennlp.apache.org/index.html}} tokenizer and sentence delimiter and our own POS tagger\footnote{\footnotesize \url{https://github.com/rasoolims/SemiSupervisedPosTagger}}. Thus the user has to download a specific sentence boundary detection and word tokenizer model from OpenNLP website depending on the specific target language. It is also possible to train a new sentence boundary detection and word tokenizer model with OpenNLP\footnote{For more information please visit OpenNLP manual at \footnotesize \url{https://opennlp.apache.org/documentation/1.5.3/manual/opennlp.html}.}.

The number of threads can be changed via the option \texttt{\footnotesize nt:[\#nt]} (e.g. \texttt{\footnotesize nt:10}). The pipeline can be downloaded from {\footnotesize \url{https://github.com/rasoolims/YaraPipeline}}.

\texttt{\footnotesize \\>> java -jar jar/YaraPipeline.jar -input [input file] -output [output file] -parse\_model [parse model file] -pos\_model [pos model] -tokenizer\_model [tokenizer model] -sentence\_model [sentence detector model]\\}

\subsection{Pipeline API usage}
It is possible to use the Yara API directly\footnote{\url{https://github.com/yahoo/YaraParser/blob/master/src/YaraParser/Parser/API_UsageExample.java}}, but the pipeline gives an easier way to do it with different levels of information. The user can set the number of threads for parsing: \texttt{\footnotesize numberOfThreads}.

\subsubsection{Importing libraries}
The user should first import libraries into the code as in Listing \ref{lst:import}. Class \texttt{\footnotesize  YaraPipeline.java} contains static methods for parsing a sentence, and  \texttt{\footnotesize ParseResult}
 contains information about words, POS tags, dependency labels and heads, and normalized tagging score and parsing score. \texttt{\footnotesize Info} contains all information about parsing setting and models for the parser, POS tagger, tokenizer and sentence boundary detector. 
 
\begin{lstlisting}[caption={Code for importing necessary libraries.},label=lst:import]
import edu.columbia.cs.rasooli.YaraPipeline.Structs.Info;
import edu.columbia.cs.rasooli.YaraPipeline.Structs.ParseResult;
import edu.columbia.cs.rasooli.YaraPipeline.YaraPipeline;
\end{lstlisting}

\subsubsection{Parsing Raw Text File}
In this case, we need to have all models for parsing, tagging, tokenization and sentence boundary detection. Listing \ref{lst:case1} shows such a case where the parser puts the results in CoNLL format into the \texttt{\footnotesize [output\_file]}.

\begin{lstlisting}[caption={Code for parsing raw text file},label=lst:case1]
// should put real file path in the brackets (e.g. [parse_model])
Info info1=new Info("[parse_model]","[pos_model]","[tokenizer_model]","[sentence_model]", numberOfThreads);
YaraPipeline.parseFile("[input_file]","[output_file]",info1);
\end{lstlisting}

\subsubsection{Parsing Raw Text} Similar to parsing a file, we can parse raw texts. It is shown in Listing \ref{lst:case2}.

\begin{lstlisting}[caption={Code for parsing raw text},label=lst:case2]
// should put real file path in the brackets (e.g. [parse_model])
Info info2=new Info("[parse_model]","[pos_model]","[tokenizer_model]","[sentence_model]", numberOfThreads);
String someText="some text....";
String conllOutputText2= YaraPipeline.parseText(someText,info1);
\end{lstlisting}

\subsubsection{Parsing a Sentence} For the cases where the user uses his own sentence delimiter, it is possible to parse sentences as shown in Listing \ref{lst:case3}.

\begin{lstlisting}[caption={Code for parsing a sentence},label=lst:case3]
// should put real file path in the brackets (e.g. [parse_model])
Info info3=new Info("[parse_model]","[pos_model]","[tokenizer_model]", numberOfThreads);
String someSentence="some sentence.";
ParseResult parseResult3= YaraPipeline.parseSentence(someSentence, info1);
String conllOutputText3=parseResult3.getConllOutput();
\end{lstlisting}

\subsubsection{Parsing a Tokenized Sentence} Listing \ref{lst:case4} shows an example for the cases where the user only wants to use the parser and POS tagger to parse a pre-tokenized sentence.

\begin{lstlisting}[caption={Code for parsing a tokenized sentence},label=lst:case4]
// should put real file path in the brackets (e.g. [parse_model])
Info info4=new Info("[parse_model]","[pos_model]", numberOfThreads);
String[] someWords4={"some", "words","..."};
ParseResult parseResult4= YaraPipeline.parseTokenizedSentence(someWords4, info1);
String conllOutputText4=parseResult4.getConllOutput();
\end{lstlisting}

\subsubsection{Parsing a Tagged Sentence}  Listing \ref{lst:case5} shows an example for the cases where the user only wants to use Yara to parse pre-tagged sentence.

\begin{lstlisting}[caption={Code for parsing a tagged sentence},label=lst:case5]
// should put real file path in the brackets (e.g. [parse_model])
Info info5=new Info("[parse_model]", numberOfThreads);
String[] someWords5={"some", "words","..."};
String[] someTags5={"tag1", "tag2","tag3"};
ParseResult parseResult5= YaraPipeline.parseTaggedSentence(someWords5,someTags5, info1);
String conllOutputText5=parseResult5.getConllOutput();
\end{lstlisting}

\section{Yara Technical Details \label{sec:yara}}
Yara is a transition-based dependency parser based on the arc-eager algorithm \cite{nivre:2004:IncrementalParsing}. It uses beam search training and decoding \cite{zhang-clark:2008:EMNLP} in order to avoid local errors in parser decisions. The features of the parser are roughly the same as \cite{zhang-nivre:2011:ACL-HLT2011} with additional Brown clustering \cite{brown1992class} features.\footnote{The idea of using Brown clustering features is inspired from \cite{koo-carreras-collins:2008:ACLMain,honnibalnon}.} Yara also includes several flexible parameters and options to allow users to easily tune it depending on the language and task. Generally speaking, there are 128 possible combinations of the settings in addition to tuning the number of iterations, Brown clustering features and beam width.\footnote{We put the best performing setting as the default setting for Yara.}

\subsection{Arc-Eager Algorithm}
As in the arc-eager algorithm, Yara has the following actions:

\begin{itemize}
\item {\bf Left-arc (LA)}: The first word in the buffer becomes the head of the top word in the stack. The top word is popped after this action.
\item {\bf Right-arc (RA)}: The top word in the stack becomes the head of the first word in the buffer. 
\item {\bf Reduce (R)}: The top word in the stack is popped. 
\item {\bf Shift (SH)}: The first word in the buffer is pushed to the stack.
\end{itemize}

\noindent Depending on position of the root, the constraints for initialization and actions differ. Figure \ref{non-monot} shows the transitions
used to parse the sentence "I want to parse a sentence .".

\paragraph{Unshift Action}
The original algorithm is not guaranteed to output a tree and thus in some occasions when the root is positioned in the beginning of the sentence, the parser decides to connect all remaining words in the stack to the root token. In \cite{nivre2014arc}, a new action and empty flag is introduced to compensate for this problem and preserve the tree constraint. The action is called \textit{unshift} which pops the first word in the stack and returns it to the start position of the buffer. We also added the ``unshift'' action for the cases where the {\em root} token is in the initial position of the sentence. This makes the parser more robust and gives a slight boost in performance.\footnote{This problem happens less in the case of beam search and it is more often in greedy parsing.}

\begin{figure*}[!t]

\centering
\centering
\small
\begin{tabular}{ *4l }

{\bf Act.} & {\bf Stack} & {\bf Buffer} & {\bf Arc(h,d)} \\   \hline 
Shift &  []& [I$_1$, want$_2$, to$_3$, parse$_4$, a$_5$, sentence$_6$, .$_7$, $\textcolor{gray}{ROOT}_8$] & \\
Left-Arc(nsubj) & [I$_1$] & [want$_2$, to$_3$, parse$_4$, a$_5$, sentence$_6$, .$_7$, $\textcolor{gray}{ROOT}_8$] & nsubj(2,1) \\
Shift & [] & [want$_2$, to$_3$, parse$_4$, a$_5$, sentence$_6$, .$_7$, $\textcolor{gray}{ROOT}_8$] &  \\
Shift & [want$_2$] & [to$_3$, parse$_4$, a$_5$, sentence$_6$, .$_7$, $\textcolor{gray}{ROOT}_8$] &  \\
Left-arc(aux) & [want$_2$, to$_3$] & [parse$_4$, a$_5$, sentence$_6$, .$_7$, $\textcolor{gray}{ROOT}_8$] & aux(4,3) \\
Right-arc(xcomp) & [want$_2$] & [parse$_4$, a$_5$, sentence$_6$, .$_7$, $\textcolor{gray}{ROOT}_8$] & xcomp(2,4) \\
Shift & [want$_2$, parse$_4$] & [a$_5$, sentence$_6$, .$_7$, $\textcolor{gray}{ROOT}_8$] & \\
Left-arc(det) & [want$_2$, parse$_4$, a$_5$] & [sentence$_6$, .$_7$, $\textcolor{gray}{ROOT}_8$] & det(6,5) \\
Right-arc(dobj) & [want$_2$, parse$_4$] & [sentence$_6$, .$_7$, $\textcolor{gray}{ROOT}_8$] & dobj(4,6) \\
Reduce & [want$_2$, parse$_4$, sentence$_6$] & [.$_7$, $\textcolor{gray}{ROOT}_8$] &  \\
Reduce & [want$_2$, parse$_4$] & [.$_7$, $\textcolor{gray}{ROOT}_8$] &  \\
Right-arc(punct)  & [want$_2$] & [.$_7$, $\textcolor{gray}{ROOT}_8$] & punct(2,7)  \\
Reduce & [want$_2$, .$_7$] & [$\textcolor{gray}{ROOT}_8$] &  \\
Left-arc(root) & [want$_2$] & [$\textcolor{gray}{ROOT}_8$] & root(8,2)  \\
DONE! &  & [$\textcolor{gray}{ROOT}_8$] &  \\
\hline 
\end{tabular}

\caption{\label{non-monot} A sample action sequence with arc-eager actions for the dependency tree in Figure \ref{fig_partial_tree}.}
\end{figure*}

\subsection{Online Learning}
Most current supervised parsers use online learning algorithms. Online learners are fast, efficient and very accurate. We use averaged structured perceptron \cite{collins:2002:EMNLP02} which is also used in previous similar parsers \cite{zhang-clark:2008:EMNLP,zhang-nivre:2011:ACL-HLT2011,choi-palmer:2011:ACL-HLT2011}. We use different engineering methods to speed up the parser, such as the averaging trick introduced by \cite[Figure 2.3]{daume06thesis}. Furthermore, all the features except label set-lexical pair features are converted to long integer values to prevent frequent hash collisions and decrease memory consumption.  Semi-sparse weight vectors are used for additional speed up, though it comes with an increase in memory consumption. The details of this implementation are out of the scope of this report. 

\subsection{Beam Search and Update Methods}
Early transition-based parsers such as the Malt parser \cite{nivre2006maltparser} were greedy and trained in batch mode. This was done by converting each tree to a set of independent actions. This has been shown to be less effective than a global search. Given our feature setting, it is impossible to use dynamic programming to get the exact result. We instead use beam search as an approximation.\footnote{Greedy search can be viewed as beam search with a beam size of one.} Therefore, unlike batch learning, the same procedure is used for training and decoding the parser. Yara supports beam search and its default beam size is 64. 

There are several ways to update the classifier weights with beam learning. A very trivial way is to get the best scoring result from beam search as the prediction and update the weights compared to the gold. This is known as ``late update'' but it does not lead to a good performance \cite{huang-fayong-guo:2012:NAACL-HLT}. A more appealing way is to keep searching until the gold prediction goes out of the beam or the search reaches the end state. This is known as "early update" \cite{collins-roark:2004:ACL} and studies have shown a boost in performance relative to late update \cite{collins-roark:2004:ACL,zhang-clark:2008:EMNLP}. The main problem with early update is that it does not update the weights according to the maximally violated prediction. A "max-violation" is a state in the beam where the gold standard is out of the beam and the gap in the score of the gold prediction and best scoring beam item is maximum. With max-violation update \cite{huang-fayong-guo:2012:NAACL-HLT}, the learner updates the weights according to the max-violation state. In other words, max-violation is the worst mistake that the classifier makes in the beam compared to the gold action. Yara supports both early and max-violation update while Zpar \cite{zhang-nivre:2011:ACL-HLT2011} only supports early update and RedShift \cite{honnibalnon} only supports max-violation. Its default value for the update model is max-violation.

\subsection{Dynamic and Static Oracles}
With the standard transition-based parsing algorithms, it is possible to have a parse tree with different action sequences. In other words, different search paths may lead to the same parse tree.
Most of the off-the-shelf parsers such as Zpar \cite{zhang-nivre:2011:ACL-HLT2011} define some manual rules for recovering a gold oracle to give it to the learner. This is known as a static oracle. The other way is to allow the oracle to be dynamic and let the learner choose from the oracles \cite{goldberg2013training}. Yara supports both static and dynamic oracles. In the case of dynamic oracles, only zero-cost explorations are allowed. In \cite{goldberg2013training}, the gold oracle can be chosen randomly but we also provided another option to choose the best scoring oracle as the selected oracle. The latter way is known as latent structured Perceptron \cite{sun2013latent} by supposing the gold tree as the structure and each oracle as a latent path for reaching the final structure. Our experiments show that using the highest scoring oracle gives slightly better results and thus we let it be the default option in the parser training.

\subsection{Other Properties}
\paragraph{Root Position}
In \cite{ballesteros2013going}, it is shown that the position of the root token has a significant effect on the parser performance. We allow the root to be either in the initial or final position in the sentence. The final position is the default option for Yara parser. 

\paragraph{Features}
We use roughly the same feature set as \cite{zhang-nivre:2011:ACL-HLT2011}. The extended feature set is the default but the user can use the \textit{\footnotesize basic} option to set it to basic set of local features to improve speed with a loss in accuracy. We also add extra features from Brown word clusters \cite{brown1992class}, as used in \cite{koo-carreras-collins:2008:ACLMain}, by using the Brown clusters for the first word in the buffer and stack, the prefixes of length 4 and 6 from the cluster bit string in the place of part of speech tags and the full bit string of the cluster in the place of words. When using all the features, we get a boost in performance but at the expense of speed. 

\paragraph{Unlabeled Parsing}
Although the parser is designed for labeled parsing, unlabeled parsing is also available through command line options. This is useful for the cases where the user simply needs a very fast parser and does not care about the loss in performance or the lack of label information.

\paragraph{Partial Parsing}
There are some occasions especially in semi-supervised parsing, where we have partial information about the tree, for example, we know that the third word is the subject of the first word. With partial parsing, we let the user benefit from dynamic oracles to parse partial trees such that known arcs are preserved unless the tree constraints cannot be satisfied. 

\paragraph{Multithreading}
Given the fact that current systems have multiple processing unit cores and many of those cores, support hyper-threading, we added the support for multithreading. When dealing with a file, the parser does multithreaded parsing on the sentence level (i.e. parsing sentences in parallel but outputting them in the same order given in the input). When using the API, it is possible to use multithreading at the beam-level. Beam level multithreading is slower than sentence-level multithreading. We also use beam-level multi-threading for training the parser and this significantly speeds up the training phase. Yara's default is set to 8 threads but the user can easily change it.

\paragraph{Model Selection}
Unlike most current parsers, Yara saves the model file for all training iterations and lets the user choose the best performing model based on the performance on the development data. It also reports the performance on the development data to make it easier for the users to select the best model.

\paragraph{Tree Scoring}
Yara also has the option to output the parse score to a text file. The score is the perceptron score divided by the sentence length. 

\paragraph{Lowercasing} In cases, such as spoken lanuage parsing, no casing is provided and it is better to train on lowercased text. Yara has this option with the argument \textit{\footnotesize lowercase} in training.

\section{Experiments\label{sec:experiments}}
In this section we show how Yara performs on two different
data sets and compare its performance to other leading
parsers.  We also graphically depict the tradeoff between
beam width and accuracy and number of iterations.  For all
experiments we use version 0.2 of Yara. We use a multi-core 2.00GHz Intell Xeon machine. The machine has twenty cores but we only use 8 threads (parser's default) in all experiments.

\subsection{Parsing WSJ Data}
 We use the the traditional WSJ train-dev-test split for our experiment. As in \cite{zhang-nivre:2011:ACL-HLT2011}, we first converted the WSJ data \cite{marcus1993building} with Penn2Malt\footnote{\footnotesize\url{stp.lingfil.uu.se/~nivre/research/Penn2Malt.html}}.
Next, automatic POS tags are generated for the whole dataset with version 0.2 of our POS tagger\footnote{\footnotesize\url{https://github.com/rasoolims/SemiSupervisedPosTagger}.} by doing 10-way jack-knifing on the training data.  The tagger is a 20-beam third-order tagger trained with the maximum violation strategy with the same settings as in \cite{collins:2002:EMNLP02}, along with additional Brown clustering features \cite{liang2005semi}.\footnote{We use the pre-built Brown cluster features in \url{http://metaoptimize.com/projects/wordreprs/} with 1000 word classes.}  
It achieved a POS tagging accuracy of 97.14, 97.18 and 97.37 on the train, development and test files respectively.

Table \ref{tab:tab_wsj1} shows the results on WSJ data by 
varying beam size and the use of Brown clusters.  A comparison 
with prior art is made in Table \ref{tab:tab_wsj2}.  All unlabeled
accuracy scores (UAS) and labeled accuracy scores (LAS) are 
calculated with punctuations ignored.
As seen in Table \ref{tab:tab_wsj1}, Yara's accuracy is very close to the state-of-the-art \cite{bohnet-nivre:2012:EMNLP-CoNLL}.

\begin{table}[t!]
\centering
\begin{tabular}{|l|c|l | c| c| c |c| c| c| }
\hline
 Parser & beam	& Features	& Iter\# &	Dev UAS & 	Test UAS &	Test LAS & Sent/Sec \\ \hline
 Yara &	1 &	ZN (basic+unlabeled)  &	 5 & 89.29	 & 88.73	 &	-- & 3929 \\
Yara &	1 &	ZN (basic)  &	6 &	89.54 &	89.34 &	88.02 & 3921 \\
Yara &	1 &	ZN  + BC &	13 &	89.98 &	89.74 &	88.52 & 1300\\
Yara &	64 &	ZN &	13 &	93.31 &	92.97 &	91.93 & 133\\
Yara &	64 &	ZN + BC &	13 &	{\bf 93.42} &	{\bf 93.32} &{\bf 92.32}  & 45 \\\hline
\end{tabular}
\caption{Parsing accuracies of Yara parser on WSJ data. BC stands for Brown cluster features, UAS for unlabeled attachment score, LAS for labeled attachment score and ZN for \cite{zhang-nivre:2011:ACL-HLT2011}.  Sent/sec refers to the 
speed in sentences per second.  }
\label{tab:tab_wsj1}
\end{table}

\begin{table}[ht!]
\centering
\begin{tabular}{ | c | l | l | c | c |   }
\hline
  & Parser & 	UAS &	LAS    \\ \hline
\multirow{ 10 }{*}{\rotatebox{90}{Graph-based}} &
 \cite{McDonald:2005:NDP:1220575.1220641} & 90.9 & --  \\
 & \cite{mcdonald2006online} & 91.5 & --  \\
 & \cite{Sagae:2006:PCR:1614049.1614082} & 92.7 & -- \\
 & \cite{koo-collins:2010:ACL} & 93.04 & -- \\
 & \cite{zhang-mcdonald:2012:EMNLP-CoNLL} & 93.06 & -- \\
& \cite{martins-almeida-smith:2013:Short} & 93.07 & -- \\
& \cite{qian2013branch} & 93.17 & --  \\
 & \cite{ma-zhao:2012:POSTERS} & 93.4 & --  \\
&  \cite{zhang-EtAl:2013:EMNLP2} & 93.50& 92.41 \\
& \cite{zhang-mcdonald:2014:P14-2} & {\bf 93.82} & {\bf 92.74}  \\
 \hline
 \multirow{ 7 }{*}{\rotatebox{90}{Transition-based}} &
 \cite{nivre2006maltparser} & 88.1 & 86.3 \\
& \cite{zhang-clark:2008:EMNLP} & 92.1 & --  \\
&  \cite{huang-sagae:2010:ACL} & 92.1 & --  \\
& \cite{zhang-nivre:2011:ACL-HLT2011} &	92.9 &	91.8  \\
& \cite{bohnet-nivre:2012:EMNLP-CoNLL} & {\bf 93.38} & {\bf 92.44}  \\
& \cite{choi-mccallum:2013:ACL2013} &  92.96 & 91.93  \\ 
\cline{2-4}
& {\bf Yara}  &	{\bf 93.32} &{\bf 92.32} \\\hline
\end{tabular}
\caption{Parsing accuracies on WSJ data. We only report
results which use the standard train-dev-test splits and
do not make use of additional training data (as in self-training). The first block of rows are the graph-based parsers and the second
block are the transition-based parsers (including Yara).}
\label{tab:tab_wsj2}
\end{table}

\begin{figure}[t!]
\centering

\includegraphics[width=0.9\textwidth,height=0.6\textwidth]{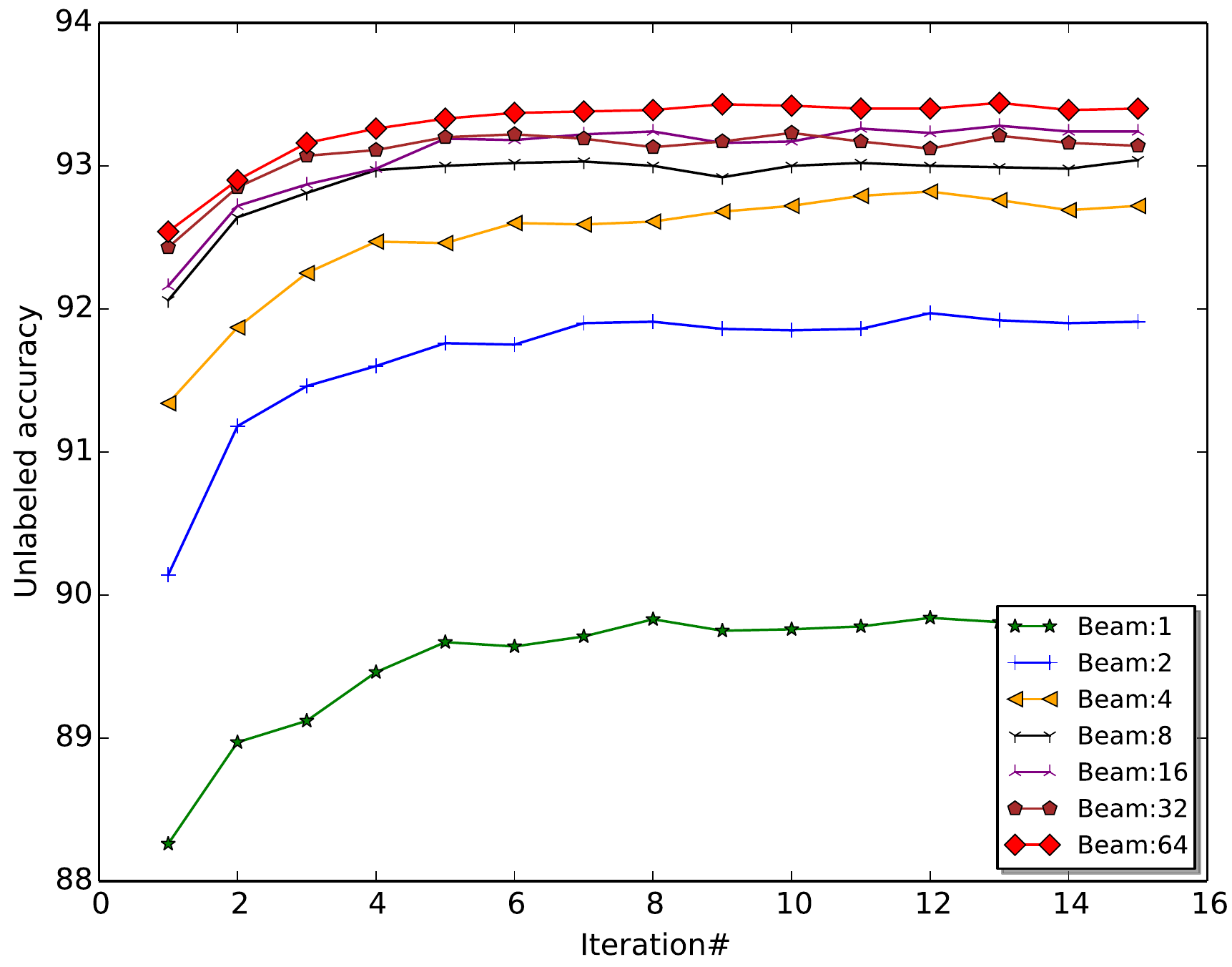}
\caption{The influence of beam size on each training iterations for Yara parser. Yara is trained with Brown clusters in all of the experiments in this figure.}
\label{fig:trend}
\end{figure}

\paragraph{Effect of Beam Size}
Choosing a reasonable beam size is essential in certain NLP applications as there is always a trade-off between speed and performance. As shown in Figure \ref{fig:trend}, after a beam size of eight, the performance results do not change as much as the performance gap in for example beam of size one compared to beam of size two. This is useful because when changing the beam size from 64 to 8, one may
speed up parsing by a factor of three (as shown in Table \ref{tab:speed_tradeoff})  with a 
small relative loss in performance.

\begin{table}[ht!]
\centering
\begin{tabular}{|l | c c | c c c c c c c |}
\hline
 Beam Size & 1 (ub) & 1 (b) & 1 & 2 & 4 & 8 & 16 & 32 & 64 \\ \hline
Dev UAS & 89.29 & 89.54 & 89.98 & 91.95 & 92.80 & 93.03 & 93.27 & 93.22 & 93.42  \\ 
Speed (sen/sec) & 3929 & 3921 & 1300 & 370 & 280 & 167 & 110 & 105 & 45  \\ \hline
\end{tabular}
\caption{Speed vs. performance trade-off when using Brown clustering features and parsing CoNLL files with eight threads (except 1 (ub) and and 1(b) which are unlabeled and labeled parsing with basic features).  The numbers are averaged over 20 training iterations and parsing development set after each iteration.}
\label{tab:speed_tradeoff}
\end{table}

\begin{table}
\centering
\begin{tabular}{|l | c c | }
 \hline
{\bf Model} & {\bf Unlabeled accuracy} & {\bf Labeled accuracy } \\ \hline
Mate v3.6.1 & 91.32 &  87.68 \\ \hline
Yara (without Brown clusters) & 89.52  &  85.77 \\ 
Yara (with Brown clusters) &  89.97 &  86.32 \\ \hline
\end{tabular}
\caption{Parsing results on the Persian treebank excluding punctuations}
\label{tab:persian}
\end{table}

\subsection{Parsing Non-Projective Languages: Persian}
As mentioned before, Yara can only be trained on projective trees and thus there will be some loss in accuracy for non-projective languages. We use version 1.1 of the Persian dependency treebank (PerDT) \cite{rasooli-kouhestani-moloodi:2013:NAACL-HLT}\footnote{\footnotesize\url{http://www.dadegan.ir/catalog/perdt}} and tagged it with the same setting as WSJ data. We tokenized Mizan corpus\footnote{\footnotesize\url{http://www.dadegan.ir/catalog/mizan}} and add it to our training data to create 1000 Brown clusters.\footnote{The definition of Brown cluster in this data is loose because there are multi-word verbs in the treebank while Brown clusters are acquired from training on single words. Therefore multi-word verbs in the treebank will not get any Brown cluster assignment and thus we will have a slight loss in performance.} The training data contains 22\% non-projective trees. We use Mate parser (v3.6.1) \cite{Bohnet:2010:VHA:1873781.1873792} as a highly accurate non-projective parsing tool to compare with Yara. Table \ref{tab:persian} shows the performance for the two parsers. There is a 1.35\% gap in unlabeled accuracy but given that 22\% of the trees ($\sim$2.5\% of the arcs) are non-projective, this gap is reasonable.

\section{Conclusion and Future Work\label{sec:conclusion}}
We presented an introduction to our open-source dependency parser. We showed that the parser is very fast and accurate. This parser can also be used for non-projective languages with a very slight loss in performance. We believe that our parser can be useful in different downstream tasks given its performance and flexible license. Our future plans include extending this parser to handle non-projectivity and also use continuous value representation features such as word embeddings to improve the accuracy of the parser.

\section*{Acknowledgements}

We would like to thank Yahoo labs open-sourcing team to allow us to release the parser with a very flexible license, especially Dev Glass for setting up the initial release of the code. We thank Amanda Stent, Kapil Thadani, Idan Szpektor and Yuval Pinter and other colleagues in Yahoo labs for their support and fruitful ideas. Finally we thank Michael Collins and Matthew Honnibal for their feedback.

\bibliographystyle{apalike} 
 \bibliography{refs}
\end{document}